\def\year{2019}\relax
\documentclass[letterpaper]{article} 
\usepackage{aaai19}  
\usepackage{times}  
\usepackage{helvet}  
\usepackage{courier}  
\usepackage{url}  
\usepackage{graphicx}  
\usepackage{subcaption}

\begin{document}
\title{Human Motion Prediction via Learning Local Structure Representations and Temporal Dependencies}
\author{Xiao Guo, Jongmoo Choi\\
University of Southern California, Department of Computer Science\\
\{xiaoguo, jongmooc\}@usc.edu\\
}
\maketitle
\begin{abstract} Human motion prediction from motion capture data is a classical problem in the computer vision, and conventional methods take the holistic human body as input. These methods ignore the fact that, in various human activities, different body components (limbs and the torso) have distinctive characteristics in terms of the moving pattern. In this paper, we argue local representations on different body components should be learned separately and, based on such idea, propose a network, Skeleton Network (SkelNet), for long-term human motion prediction. Specifically, at each time-step, local structure representations of input (human body) are obtained via SkelNet's branches of component-specific layers, then the shared layer uses local spatial representations to predict the future human pose. Our SkelNet is the first to use local structure representations for predicting the human motion. Then, for short-term human motion prediction, we propose the second network, named as Skeleton Temporal Network (Skel-TNet). Skel-TNet consists of three components: SkelNet and a Recurrent Neural Network, they have advantages in learning spatial and temporal dependencies for predicting human motion, respectively; a feed-forward network that outputs the final estimation. Our methods achieve promising results on the Human3.6M dataset and the CMU motion capture dataset. 

\end{abstract}
\section{Introduction} 
\begin{figure}[t]
\includegraphics[width=8.5cm]{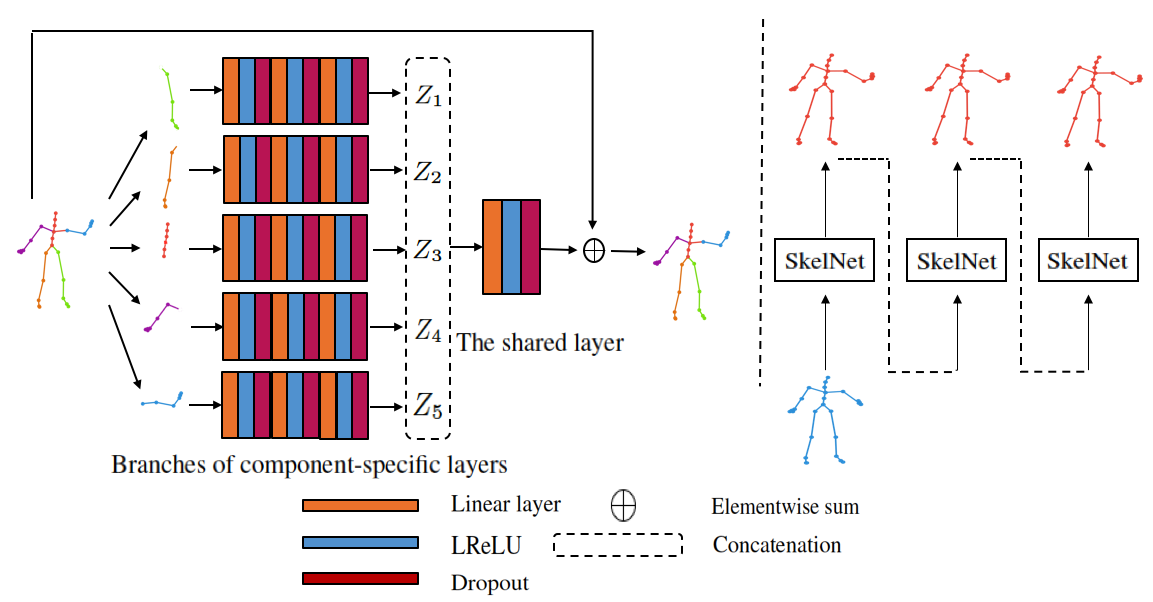}
\centering
\caption{\textit{Left}: An illustrative sketch of proposed SkelNet. The human pose is divided into five non-overlapping parts depicted in different colors, which are fed accordingly into five branches of component-specific layers for learning local structure representations. $z_1, z_2,  z_3, z_4, z_5$ respectively denote representations for different body components. The shared layer uses representations of local structure to predict the future human pose. \textit{Right}: During both training and testing, SkelNet is fed with one seed mocap vector, then predicts the mocap sequence by always sampling its own generated samples.}
\label{fig:SkelNet}
\end{figure}
Human motion prediction from motion capture (mocap) data is attracting a significant attention in recent years, which has been applied in a variety of fields: human pose tracking \cite{taylor2010dynamical,Tekin_2016_CVPR}, Physics-Based Motion synthesis \cite{liu2005learning,arikan2003motion} and human-computer interaction \cite{koppula2016anticipating}. However, estimating physical parameters, like skeleton joint angles, used to represent human poses from mocap data is still a challenging issue, as the realistic human motion is involved with complexities in kinematics and variations of the dynamic pattern, which are difficult to predict.\\
\indent Our goal is to forecast future human poses from mocap data via proposed networks, in which \textit{each mocap feature vector is a set of skeleton joint angles of the human pose.} In fact, our prediction task can be divided into two subtasks: long-term and short-term predictions. In long-term scenario, errors can accumulate severely with long time horizons; in short-term, motions are more certain and constrained by the temporal coherence. Many previous methods 
\cite{taylor2007modeling,taylor2010dynamical,Fragkiadaki_2015_ICCV,Jain_2016_CVPR,Martinez_2017_CVPR,Butepage_2017_CVPR} have achieved promising results on these two subtasks by using deep learning approaches. However, we observe that different physical components (limbs and the torso) of the human body participate into actions to varying degrees. These components have different moving dynamics and should be processed separately, but the previous work is fed with the human pose that is never fully partitioned in a component-specific way, this violates the kinematics principle of the human body, and may bring in intra-pose interventions in generating future human poses.\\
\indent In this paper, we propose a novel network at first for long-term human motion prediction, named as Skeleton Network (SkelNet). Unlike the previous work ignoring distinctive characteristics in terms of the moving pattern of different human body components, SkelNet uses branches of component-specific layers for learning local structure representations of different human body components. Our SkelNet (Figure \ref{fig:SkelNet}) is a variant to the standard feed-forward network, after being fed with the seed mocap feature vector(s), it generates a mocap sequence by always sampling previous self-generated samples. Specifically, instead of taking the holistic human pose as input, we divide it into five non-overlapping parts according to the human physical structure, and then separately feed these five parts into SkelNet's five branches of component-specific layers. SkelNet's component-specific layers learn local spatial representations of input at the current time-step, and a shared layer is responsible to use these local representations for estimating the mocap feature vector at the next time-step. To the best of our knowledge, SkelNet is the first to predict the human motion via local spatial representations of the human pose, where its branch structure learns component-independent information from different human body-part domains. Moreover, SkelNet also derives its generalization capability from a special network design, which is based on a set of simple ideas, such as adding the residual connection, using dropout and the nonlinear activation function.\\
\indent Furthermore, temporal dependencies interpret temporal coherence imposed by the human motion over time horizons, which is important for predicting future human poses, especially when motion is more certain within a short time range. Hence, we propose the second network, Skeleton Temporal Network (Skel-TNet), for short-term prediction task, it employs SkelNet in the company of Recurrent Neural Network (RNN) that is known as an effective technique for modeling temporal dynamics. In fact, our Skel-TNet has three components (Figure \ref{Skel-TNet}), aside from SkelNet, we use a RNN with one Gate Recurrent Unit (GRU) \cite{cho2014learning}, named as C-RNN, which can efficiently learn temporal dynamics of the human pose sequence. The third component is Merging Network, an another feed-forward network that aims at generating the final estimated mocap sequence. Skel-TNet is a multistage processing framework, three components need to be trained separately yet an end-to-end fashion is maintained in the evaluation. It demonstrates a new way for predicting short-term human motion: to leverage results generated from pre-trained models, which respectively have advantages in learning local spatial structure representations and temporal dependencies.\\
\indent In summary, our contributions are: 1) SkelNet, a new network that uses local structure representations for predicting long-term human motion. This is the first time that local spatial representations are used for human motion prediction, 2) Skel-TNet, a new network for predicting short-term human motion, it takes advantages of capabilities on learning local structure representations and temporal dependencies from proposed SkelNet and a GRU-based RNN, respectively, 3) experimental results demonstrate that our methods achieve promising results on human motion prediction, compared to the state-of-art method, and SkelNet exhibits the meaningful robustness towards noise.
\section{Related work} 
\textbf{Human motion prediction in mocap data.} 
Many works are based on traditional statistical methods for predicting or synthesizing the human motion \cite{LGCG12,zhao2018adversarial,Wang06gaussianprocess,Wu_2014_CVPR}. Meanwhile, deep learning approaches have also achieved remarkable accomplishments. Specifically, \cite{Fragkiadaki_2015_ICCV} present two end-to-end discriminatively trained models: Encoder-Recurrent-Decoder (ERD) and a LSTM recurrent neural network (LSTM-3LR), for modeling human kinematics from videos and mocap data. In \cite{Jain_2016_CVPR}, authors cast an arbitrary, scalable and jointly trainable stacked RNN based LSTMs, called structural RNN (SRNN). Then residual RNN (RRNN) is proposed by \cite{Martinez_2017_CVPR}, which solve issue of discontinuity in initial frames by a relatively simple and intuitive way, and after that, \cite{ghosh2017learning} build model for predicting human motion in the long run. Most recently, \cite{Li_2018_CVPR} use a convolutional sequence to sequence (seq2seq) \cite{sutskever2014sequence} model (CSS) for learning both global spatial dependencies and long-term temporal dependencies. It is noteworthy that one common feature of some existing methods including \cite{Butepage_2017_CVPR} is their heavy reliance on long prior information encoded from many previous observed mocap frames, in contrast, our SkelNet does not need such help, still being able to produce encouraging results. \\
\textbf{Architecture with Multi-stage processing.} Several previous methods construct frameworks which are multi-stage processing, or combine multiple methods that are streamlined processing, for obtaining promising results on various tasks: object detection and segmentation \cite{Girshick_2014_CVPR}; face hallucination \cite{Yu_2017_CVPR}; various tasks in image synthesis \cite{Huang_2017_CVPR,Han17stackgan2}. In short, networks operating in multi-stages can benefit from being conditioned on intermediate results, as well as specific capabilities provided by pretrained models. What is more, spatial and temporal dependencies are two ideas that have been applied in a variety of works relating to the human analysis: human motion tracking \cite{xu2018geometrical}, 3D pose estimation \cite{fang2017learning}, activity recognition \cite{Liu_2017_CVPR,liu2016spatio} and robotics \cite{koppula2016anticipating}, etc. Our Skel-TNet uses both local spatial dependencies and temporal dependencies for predicting the human motion, this is different from the previous work who \cite{Jain_2016_CVPR} attempts to learn spatial and temporal dependencies altogether in one single largely complicated spatio-temporal graph, and who \cite{ghosh2017learning} obtains the mediocre result by training two independent networks jointly. \\
\section{Proposed method}
\subsection{SkelNet}
\begin{figure}[t]
\includegraphics[width=8.5cm]{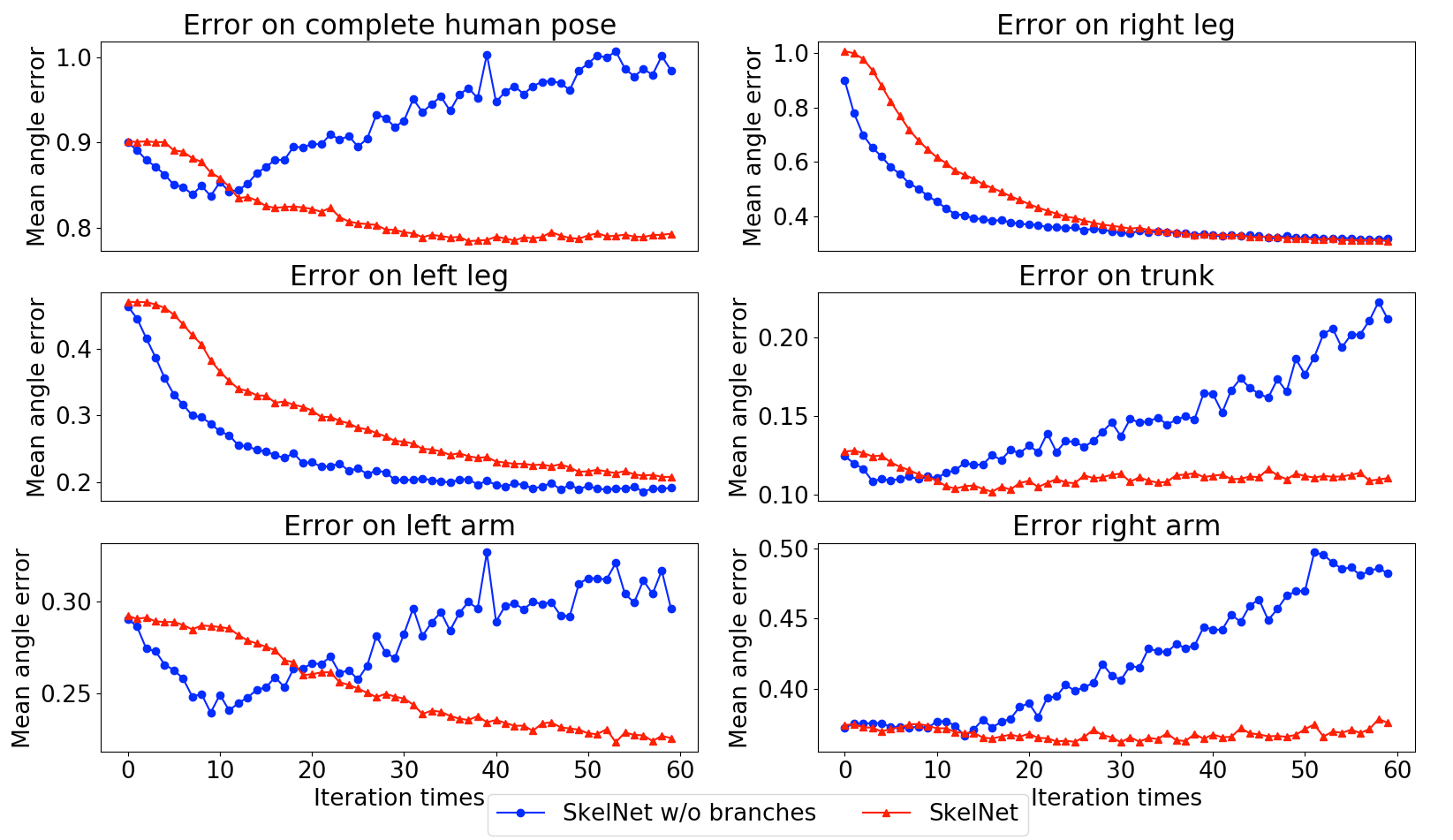}
\caption{ Six plots record Euler angle errors on the complete human pose and their five different human component groups, respectively. On predicting long-term “Eating” activity in the Human3.6M dataset, Red lines indicate errors from the SkelNet and Blue lines indicate error from the SkelNet without branches.}
\label{Skel_Plain}
\end{figure}

\textbf{Architecture:} As previously stated, many previous methods rely on using large number of observed frames as the prior information, but observed frames can be unavailable and corrupted with noise, so we choose a standard feed-forward network as the the basic sequence generator (baseline) instead of the seq2seq model, it outputs the mocap vector $y_t (t\geq 1)$, as estimation to the one-step-ahead ground truth vector $x_t$, given self-generated mocap vectors (including seed one) so far (Figure \ref{fig:SkelNet}). The residual connection \cite{He_2016_CVPR} allows the network to predict the motion velocity instead of the whole human pose, this achieves encouraging results without taking considerable efforts, as demonstrated in \cite{Martinez_2017_CVPR}. We also incroporate leaky Rectified Linear Units (LReLUs) \cite{Maas2013RectifierNI}, dropout \cite{JMLR:v15:srivastava14a} into the network design for adding non-linearity and combating overfitting. Leveraging these simple ideas in the deep learning enhances the generalization capability of our baseline on human motion prediction, details are reported in Table \ref{t_skelNet_archi}.\\
\indent The first several layers of SkelNet is divided into five branches. Being different with the previous work, regrading input data at each time-step as a complete one, we divide such input into five non-overlapping parts, according to the upper left/ right (arm), the lower left/ right (leg) and the torso. These five inputs are fed into five component-specific branches accordingly, and branches learn local structure representations of the human pose. Such partitioned data altogether with the branch structure make SkelNet have the higher generalization capability than that of networks used in the previous work. This is due to the fact that different physical parts of the human pose have different level involvements in certain activities, such that five different kinematical information of the human pose should be treated and processed separately. For example, activities like ``walking'' and ``running'' exhibit a large amount of variations in lower skeletal parts (two legs), yet upper parts (arms) matter less; people wave, lift and stretch out their arms to do ``directions'', ``washing window'' and ``eating" while being probably hold legs and torso still.\\
\indent The input layer in a conventional feed-forward network takes in data without partitioning and project the high dimensional vector as output, each dimension (value) is decided jointly by the spatial information of the entire human pose, namely five assigned groups altogether. This suggests that, throughout the entire neural network, every single unit considers global information of the complete human pose, this can cause intra-pose interventions, which hinder us from learning local structural representations of the body configuration for forecasting future poses. Conversely, in SkelNet, data from different groups are passed through branches accordingly, such simple modification largely overcomes downside of intra-pose interventions, focusing information of the locality like limb-specific or torso-specific information. Figure \ref{Skel_Plain} demonstrates our idea, and more comprehensive results are reported in ablative study.\\

\textbf{Robustness analysis:} If input data is corrupted by noise,  we conjecture that the previous work will witness larger declines in the prediction accuracy than SkelNet. Evidently, it is because that some of the previous work (e.g., RRNN and CSS) uses long prior information encoded from a large number of previous mocap frames, whose guidance meaning will diminish if there are missing frames or skeleton poses corrupted by noise. On the contrary, SkelNet counts generalization capability on the particular network design, without the need of long prior information, so SkelNet is likely to make the more robust choice. However, this is partially proved. We evaluate models for the prediction task, where input is under the Gaussian noise. SkelNet outperforms previous methods on the CMU motion capture dataset but keeps comparable in the Human3.6M dataset, relating details will be reported in Table \ref{noise_resiliency}. Even so, SkelNet still exhibits meaningful robustness towards noise. \\
\indent In short-term prediction, motion is more constrained by the temporal coherence, this has been demonstrated in \cite{Fragkiadaki_2015_ICCV,Martinez_2017_CVPR} and thereby the RNN-based method is a popular approach. However, our SKelNet constructed as a lightweight architecture cannot learn temporal dynamics as effectively as RNN-based methods equipped with LSTMs and GRUs, which are known as efficient ways for modeling temporal dependencies by adopting multiple gates. To handle this issue, we design Skeleton Temporal Network (Skel-TNet) based on SkelNet.
\subsection{Skel-TNet}
\begin{figure}[t]
\includegraphics[width=9 cm]{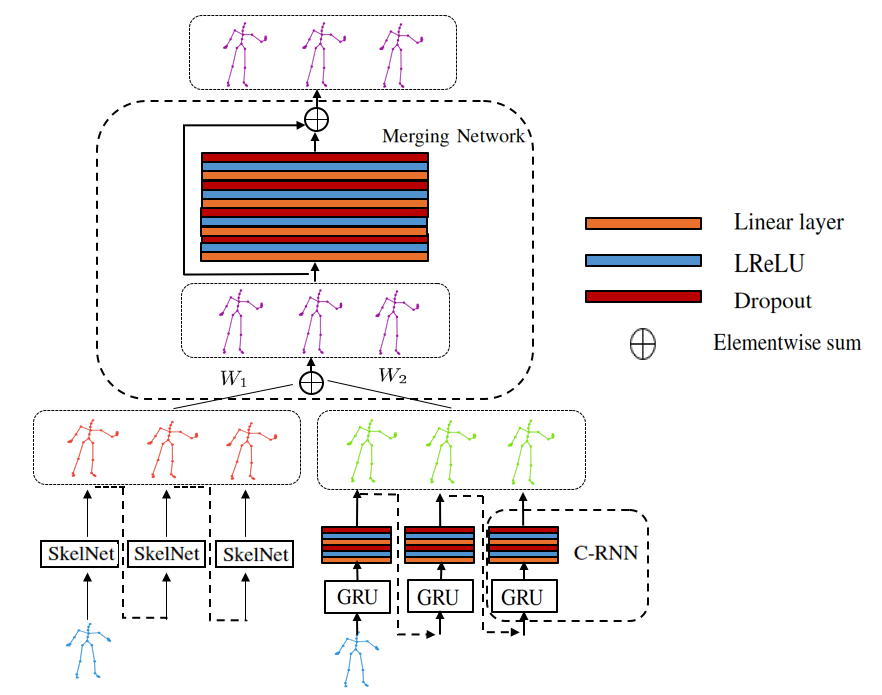}
\caption{Skel-TNet has three main components: SkelNet, C-RNN and Merging Network. We train SkelNet and C-RNN at the first stage. Pre-trained models generate two sequences (Red and Green) as inputs to Merging Network, which outputs the final generated sequence (Purple). In testing, three components are maintained in an end-to-end fashion. At each time-step, the residual connection used to bridge input and output in C-RNN are not depicted.}
\label{Skel-TNet}
\end{figure}

\textbf{Architecture:} For short-term human motion prediction, we propose Skel-TNet for predicting the human motion, in which SkelNet works with a RNN. As a matter of fact, the major motivation behind Skel-TNet is making use of different models' advantages in learning spatial and temporal dependencies, both of which play significant roles in our prediction task.\\
\indent Skel-TNet has three components (Figure \ref{Skel-TNet}). Aside from SkelNet that predicts the human motion via learning local structure representations, we use a RNN based on GRU that can efficiently model temporal dynamics, we name it as C-RNN. C-RNN has one standard GRU, as a computationally less-expensive alternative to LSTMs, followed by two linear layers with LReLUs, dropout and the residual connection. It predicts the future sequence given the seed mocap frame(s) along with self-geneated samples. Our C-RNN achieves more promising results via being trained by our proposed \textit{Converging\textunderscore loss}, which will be introduced later. SkelNet and C-RNN separately output two estimated sequences, as intermediate results, sent to the third component, which is an another feed-forward neural network, called Merging Network. Our Merging Network is responsible for the final estimated sequence and it has two trainable weights, which adaptively control how much of two intermediate results are used for generating the final prediction.\\

\begin{figure}
\includegraphics[width=6.5cm]{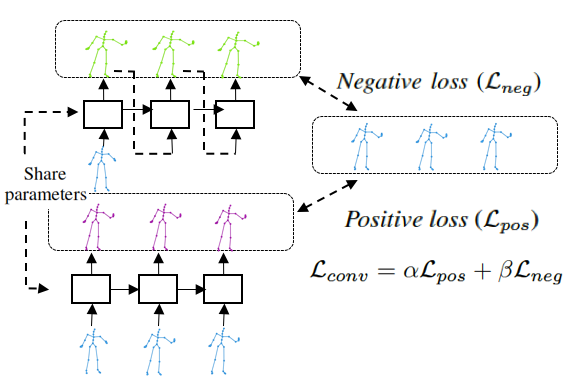}
\caption{The first output sequence (Green) is generated from the RNNs model conditioned on its own output at previous time-steps and the seed mocap frame(s); The second output sequence (Purple) is generated from the RNNs model conditioned on the ground truth at all time-steps; $\mathcal{L}_{pos}$ and $\mathcal{L}_{neg}$ is computed between them (Green and Purple sequence) and the ground truth sequence (Blue), respectively.}
\label{Conv_loss_figure}
\end{figure}

\textbf{Learning:} Optimizing RNNs by always feeding them the previous self-generated sample as input, this idea is firstly proposed by \cite{Martinez_2017_CVPR} on human motion prediction and called \textit{Sampling-based loss}. Although such training strategy improves network's ability to recover from its own mistakes along the generation, we find that it often leads to the poor human-motion prediction performance, which is because the divergence exists in training between conditioned context (the self-generated sample) and the ground truth. How to reduce the negative impact brought by this type of divergence in training, while keep updating RNNs' ability to recover from its own mistakes is the issue we need to tackle. Hence, we design \textit{Converging loss} ($\mathcal{L}_{conv}$), which is composed of two terms: \textit{Positive loss} ($\mathcal{L}_{pos}$) and \textit{Negative loss} ($\mathcal{L}_{neg}$), and $\alpha$, $\beta$ is the weight associated with $\mathcal{L}_{pos}$ and $\mathcal{L}_{neg}$. 
\begin{equation}
\mathcal{L}_{conv} = \alpha \mathcal{L}_{pos} + \beta \mathcal{L}_{neg}\label{eq:1}
\end{equation}
Specifically, these two terms are Euclidean distances between the ground truth sequence and the generated sequence, but the generated sequence used in $\mathcal{L}_{pos}$ is from RNNs always being fed by the ground truth and sequence used in $\mathcal{L}_{neg}$ is retrieved from RNNs constantly taking in self-generated samples, as shown in Figure \ref{Conv_loss_figure}. $\mathcal{L}_{conv}$ encourages RNNs to generate sequence conditioned on the ground truth in training via $\mathcal{L}_{pos}$, meanwhile forces RNNs to keep learning how to correct from its own mistakes via $\mathcal{L}_{neg}$. We empirically find that if we replace \textit{Sampling-based loss} with $\mathcal{L}_{conv}$ as the objective function in training, improvements of prediction performance on RRNN and C-RNN are evident, so we choose to train C-RNN via $\mathcal{L}_{conv}$. After optimizing SkelNet and C-RNN, we use these two pertained models to generate two mocap sequences, which are sent to Merging Network that is trained by minimizing the Euclidean distance between its estimation and the ground truth. Although three components need to be trained independently, Skel-TNet maintains the end-to-end fashion in the testing phase.

\textbf{Understanding:} For $\mathcal{L}_{conv}$, we in fact instantiate a longstanding idea on the optimization for generative RNNs: monitoring behaviors of RNNs in the ``free-running'' (negative phase) mode as well as the ``teacher forcing" (positive phase) mode \cite{williams1989learning}. Recently, one of famous instances of this idea turns out to be the professor-forcing algorithm \cite{lamb2016professor}, nevertheless \textbf{such idea has yet to be applied on the human modeling community until our $\mathcal{L}_{conv}$}. Furthermore, $\mathcal{L}_{conv}$ serves as a way of complementing information of the ground truth in the decoding phase. Moreover, chances are that $\mathcal{L}_{conv}$ could be deployed into other generation problems, such as hand-writing images reconstruction, speech generation, etc.\\
\indent Furthermore, we need to claim three points for the more solid explanation to the network design of Skel-TNet. Firstly, employing both SkelNet and C-RNN is for learning both structure representations and temporal dependencies, not for small gainings in performance, as results can be further enhanced by using the deeper feed-forward network or stacking multiple GRUs. Secondly, the main purpose of our Merging Network is designed to combine and balance two intermediate results into single one as the final output, its predictive ability is limited but which is what \cite{Butepage_2017_CVPR} mainly focuses on. This difference in motivations separates our Merging Network from their proposed network, even though it is true that these two share similar structures. Thirdly, our Skel-TNet is the first method that employs a multi-stage processing network, and each component is conceptually simple.\\
\indent In summary, Skel-TNet instantiates the idea that human motion prediction can be achieved by a multi-staged processing network, which makes use of different model's advantages, namely, SkelNet can learn local spatial structure effectively, C-RNN can learn temporal dynamics effectively.\\
\section{Experiment}
In this section, we introduce two datasets and implementation details. Then we evaluate our proposed models for several different tasks in human motion prediction. Our experiments include comparison with many previous methods, including ERD, LSTM-3LR\cite{Fragkiadaki_2015_ICCV}, SRNN\cite{Jain_2016_CVPR}, RNN\textunderscore D \cite{lin2018human}, RRNN\cite{Martinez_2017_CVPR} and CSS\cite{Li_2018_CVPR}. In the end, ablative experiments are carried out for showing each component's contribution. Our code is publicly available at link \footnote{https://github.com/CHELSEA234/SkelNet\_motion\_prediction}.

\begin{table*}[ht]
\centering
\begin{subtable}{\textwidth}
\centering
\resizebox{1\textwidth}{!}{
\begin{tabular}{ccccc ccccc ccccc cccc} 
\hline
\multicolumn{19}{c}{Human3.6M} \\
& Walk & Eat & Somke & Discuss & Direct & Greet & Phone & Pose & Purch & Sitting & SittingD & Photo & Wait & WalkD & WalkT & \multicolumn{3}{c}{Average/Standard Deviation}\\
RRNN&0.97&1.10&1.30&1.33&1.60&1.98&1.67&1.89&1.71&1.52&1.90&1.31&1.60&1.70&1.07&\multicolumn{3}{c}{1.51/\textbf{0.32}}\\
CSS&0.71&0.78&0.99&1.27&1.00&1.47&1.53&1.75&1.44&1.21&1.33&0.91&1.55&\textbf{1.45}&0.83&\multicolumn{3}{c}{1.22/\textbf{0.32}}\\
\hline
SKelNet&\textbf{0.69}&\textbf{0.77}&0.96&\textbf{1.21}&\textbf{0.96}&\textbf{1.46}&\textbf{1.50}&\textbf{1.56}&\textbf{1.41}&\textbf{1.14}&\textbf{1.24}&\textbf{0.84}&1.57&1.52&\textbf{0.78}&\multicolumn{3}{c}
{\textbf{1.17/0.32}}\\
SKel-TNet&0.70&\textbf{0.77}&\textbf{0.95}&1.22&\textbf{0.96}&1.47&1.52&1.58&1.44&1.17&1.26&0.85&\textbf{1.51}&1.52&0.80&\multicolumn{3}{c}
{1.18/\textbf{0.32}}\\
\hline
\hline
\multicolumn{19}{c}{CMU mocap dataset} \\
& \multicolumn{2}{c}{Walk} &  \multicolumn{2}{c}{Run}  &  \multicolumn{2}{c}{DirectTraffic}  & \multicolumn{2}{c}{Soccer} & \multicolumn{2}{c}{Basketball} & \multicolumn{2}{c}{WashWindow} & \multicolumn{2}{c}{Jump} & \multicolumn{2}{c}{BasketballSignal} & \multicolumn{2}{c}{Average/Standard Deviation}\\
CSS & \multicolumn{2}{c}{0.56}&\multicolumn{2}{c}{\textbf{0.48}}&\multicolumn{2}{c}{1.04}&\multicolumn{2}{c}{0.79}&\multicolumn{2}{c}{1.54}&\multicolumn{2}{c}{0.86}&\multicolumn{2}{c}{1.55}&\multicolumn{2}{c}{0.65}&\multicolumn{2}{c}{0.93/\textbf{0.42}}\\
\hline
SKelNet & \multicolumn{2}{c}{\textbf{0.51}}&\multicolumn{2}{c}{0.57}&\multicolumn{2}{c}{\textbf{0.99}}&\multicolumn{2}{c}{\textbf{0.74}}&\multicolumn{2}{c}{\textbf{1.50}}&\multicolumn{2}{c}{\textbf{0.79}}&\multicolumn{2}{c}{\textbf{1.54}}&\multicolumn{2}{c}{\textbf{0.41}}&\multicolumn{2}{c}{\textbf{0.88}/0.43}\\
SKel-TNet &\multicolumn{2}{c}{0.53}&\multicolumn{2}{c}{0.59}&\multicolumn{2}{c}{\textbf{0.99}}&\multicolumn{2}{c}{0.76}&\multicolumn{2}{c}{1.52}&\multicolumn{2}{c}{0.81}&\multicolumn{2}{c}{1.59}&\multicolumn{2}{c}{0.43}&\multicolumn{2}{c}{0.90/0.44}\\
\hline
\end{tabular}
}
\caption{Long-term human motion prediction}
\label{long_term_prediction}
\end{subtable}
\begin{subtable}{\textwidth}
\centering
\resizebox{1\textwidth}{!}{
\begin{tabular}{ccccc ccccc ccccc cccc} 
\hline
\multicolumn{19}{c}{Human3.6M} \\
& Walk & Eat & Somke & Discuss & Direct & Greet & Phone & Pose & Purch & Sitting & SittingD & Photo & Wait & WalkD & WalkT & \multicolumn{3}{c}{Average/Standard Deviation}\\
RRNN&0.51&0.61&0.77&0.91&1.06&1.20&1.28&1.31&1.08&0.94&1.11&0.82&1.02&1.04&0.75&\multicolumn{3}{c}{0.97/0.23}\\
CSS&0.49&0.44&0.63&0.73&0.65&\textbf{1.00}&\textbf{1.17}&0.95&0.92&0.76&0.83&0.57&0.82&0.99&0.53&\multicolumn{3}{c}{0.77/\textbf{0.21}}\\
\hline
SKelNet&0.49&0.46&\textbf{0.60}&\textbf{0.70}&\textbf{0.62}&1.03&1.21&0.77&0.89&0.76&\textbf{0.80}&\textbf{0.56}&0.85&1.00&0.51&\multicolumn{3}{c}{0.76/\textbf{0.21}}\\
SKel-TNet &\textbf{0.48}&\textbf{0.41}&0.61&\textbf{0.70}&\textbf{0.62}&\textbf{1.00}&1.19&\textbf{0.76}&\textbf{0.86}&\textbf{0.75}&\textbf{0.80}&0.57&\textbf{0.81}&\textbf{0.96} &\textbf{0.50}&\multicolumn{3}{c}{\textbf{0.73/0.21}}\\
\hline
\hline
\multicolumn{19}{c}{CMU mocap dataset} \\
& \multicolumn{2}{c}{Walk} &  \multicolumn{2}{c}{Run}&\multicolumn{2}{c}{DirectTraffic}  & \multicolumn{2}{c}{Soccer} & \multicolumn{2}{c}{Basketball} & \multicolumn{2}{c}{WashWindow} & \multicolumn{2}{c}{Jump} & \multicolumn{2}{c}{BasketballSignal} & \multicolumn{2}{c}{Average/Standard Deviation}\\
CSS & \multicolumn{2}{c}{0.41}&\multicolumn{2}{c}{\textbf{0.46}}&\multicolumn{2}{c}{0.60}&\multicolumn{2}{c}{0.63}&\multicolumn{2}{c}{0.86}&\multicolumn{2}{c}{0.51}&\multicolumn{2}{c}{\textbf{0.90}}&\multicolumn{2}{c}{0.58}&\multicolumn{2}{c}{0.61/\textbf{0.18}}\\
\hline
SKelNet & \multicolumn{2}{c}{\textbf{0.34}}&\multicolumn{2}{c}{0.56}&\multicolumn{2}{c}{0.57}&\multicolumn{2}{c}{0.55}&\multicolumn{2}{c}{0.84}&\multicolumn{2}{c}{0.55}&\multicolumn{2}{c}{0.96}&\multicolumn{2}{c}{0.38}&\multicolumn{2}{c}{0.60/0.21}\\
SKel-TNet & \multicolumn{2}{c}{0.36}&\multicolumn{2}{c}{0.52}&\multicolumn{2}{c}{\textbf{0.54}}&\multicolumn{2}{c}{\textbf{0.48}}&\multicolumn{2}{c}{\textbf{0.81}}&\multicolumn{2}{c}{\textbf{0.51}}&\multicolumn{2}{c}{\textbf{\textbf{0.90}}}&\multicolumn{2}{c}{\textbf{0.32}}&\multicolumn{2}{c}{\textbf{0.55}/0.21}\\
\hline
\end{tabular}
}
\caption{Short-term human motion prediction}
\label{short_term_prediction}
\end{subtable}
\caption{Long-term and short-term human motion prediction errors reported in MoF on each activity of the Human3.6M dataset and the CMU mocap dataset. Bold means the best performance.}
\end{table*}

\subsection{Experiment setup }
\textbf{Dataset and preprocessing:} Two widely used public benchmarks are chosen in our experiments: 1) Human3.6M \cite{ionescu2014human3} \nocite{ionescu2014human3}: It has 15 different activity categories performed by professional actors (subjects) from ordinary life: ``walking'', ``eating'', ``smoking'', etc. It is the largest human motion capture dataset and also has been winning increasing popularity recently. Following the previous method\cite{Martinez_2017_CVPR}, 
we use the sequences given by the subject five for testing, while rest sequences for training. The detail can be found in the public\footnote{https://github.com/una-dinosauria/human-motion-prediction}, 2) CMU motion capture (CMU mocap) dataset \cite{CmuMocap}: it contains 2235 recordings belonging to five major activity categories, being performed by 144 different subjects. This is a challenging dataset as it contains more complex activities, like ``basketball", ``soccer". Data is recorded with the mocap system and a pose are represented with 38 joints in 3D space. As with \cite{Li_2018_CVPR}, eight actions are selected for our experiments, who are already divided into as two parts for training and testing, the code is publicly available \footnote{https://github.com/chaneyddtt/Convolutional-Sequence-to-Sequence-Model-for-Human-Dynamics/tree/master/src}
. Also, two datasets are preprocessed in the same way as the previous work, in which still joints are removed and Euler angles are converted to exponential map representation, making human skeleton pose invariant to the orientation and avoid the gimbal lock effect.\\
\textbf{Implementation:} We implement our methods using the Tensorflow as the backend. Regarding SkelNet, each branch consists of three linear layers, dimensions are 64, 128, 64. The last layer is of the same dimension as input, namely dimension of each time-step’s mocap frame (54 and 70 dimensions for data in two datasets, respectively), denoted as \textit{input dimension}. We adopt the Gradient Descent optimizer with learning rate 0.01 \footnote{The choice is to make the magnitude of $\mathcal{L}_{pos}$ and $\mathcal{L}_{neg}$ are of same scale.}. In Skel-TNet, C-RNN has a GRU with 1024 units, followed by two fully-connected layers (512, \textit{input dimension}), this network is trained by $\mathcal{L}_{conv}$, where weights ahead of $\mathcal{L}_{pos}$, $\mathcal{L}_{neg}$ are set as 1 and 0.1. The learning rate is set as 5e-5 for Skel-TNet, with the gradient descent optimizer. Merging Network is as depicted in Figure \ref{Skel-TNet}, dimensions are 1024, 512, 512, \textit{input dimension}, and two input weights are initialized as 0.5. We use the adam optimizer with learning rate 0.01 to train Merging Network. Throughout the entire network, LReLUs are with the negative slope as 0.2 and dropouts are with rate 0.2 to dropout.\\
\textbf{Metric:} We adhere to the evaluation metric in the previous work: we measure Euclidean distances between our predictions and the ground truth in the angle-space for increasing time horizons, then compute the average error at each frame over eight randomly sampled test sequences. The result is reported in Table \ref{3_activities} that has three commonly used representative actions. The random seed that generates eight sequences is fixed as the same one in the previous work for the fair comparison. Also, we report mean errors over all frames along eight randomly chosen test sequences for each actions in two datasets, the result is denoted as the MoF (mean error of frames) for simplicity.
\subsection{Results and comparisons}
\textbf{Long-term motion prediction:} SkelNet is trained by minimizing the prediction error over 1000 ms in the future, and we compare it with recently proposed methods (CSS and RRNN). We report results in Table \ref{long_term_prediction}, whose rightmost column demonstrates that our results catch up with or even surpass the state-of-the-art method.  We notice that RRNN suffers from a severe overfitting issue and relatively poor prediction performances (e.g., MoF in predicting ``Walking'', ``Eating'' activities), even though it has been considered as the flagship work previously. CSS simply takes a large number of complete human poses into the network, which we believe cannot learn spatial dependencies imposed by each human skeleton pose, on the contrary, SkelNet explicitly uses branches of component-specific layers to learn the spatial dependencies. Except ``Walking-dog'' activity in the Human3.6M dataset, SkelNet outperforms two competitors in all other scenarios,  and the enhancement on predicting ``BasketballSignal'' activity is remarkable. Qualitatively, we are able to see clear movements in different motion sequences generated by SkelNet, results keep physically meaningful and realistic. In ``Walking" and ``Basketball", stick man is with footstep changed; In ``DirecTraffic", the left hand is raised; In ``Soccer", stick man kicks ball out by different body positions.\\
\indent In order to have the more concrete analysis on proposed networks' prediction performance, we also train Skel-TNet on long-term prediction task. As a result, Skel-TNet performs slightly better than CSS regarding the averaging MoF, which means it also can handle long-term prediction task.\\
\indent \\
\begin{figure}
\includegraphics[width=8.5cm]{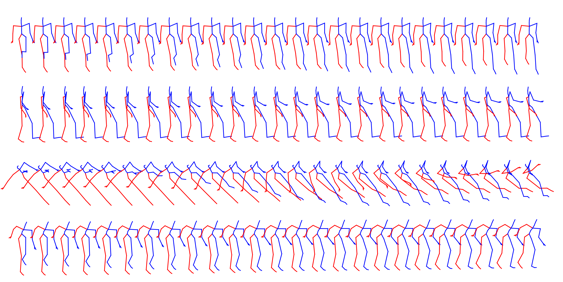}
\caption{Qualitative results on ``Walking", ``DirectTraffic", ``Soccer" and ``Basketball" activities (from top to down) in the CMU mocap dataset. }
\label{Qualitative_result}
\end{figure}
\begin{table*}[ht!]
\centering
\resizebox{0.8\textwidth}{!}{
\begin{tabular}{c| ccccc| ccccc |ccccc}
&\multicolumn{5}{c|}{Eat} &\multicolumn{5}{c|}{Smoke} &\multicolumn{5}{c}{Discuss}\\
millisecond&80&160&240&320&400&80&160&240&320&400&80&160&240&320&400\\
\hline
ERD(ICCV'15)&1.27&1.45&&1.66&1.80&166&1.95&&2.35&2.42&2.27&2.47&&2.68&2.76\\
LSTM-3LR(ICCV'15)&0.89&1.09&&1.35&1.46&1.34&1.65&&2.04&2.16&1.88&2.12&&2.25&2.23\\
SRNN(CVPR'16)&0.97&1.14&&1.35&1.46&0.97&1.14&&1.35&2.08&1.22&1.49&&1.83&1.93\\
RRNN(CVPR'17)&0.29&0.52&0.67&0.87&1.10&0.36&0.67&0.97&1.18&1.27&0.41&0.90&1.14&1.31&1.40\\
CSS(CVPR'18)&0.23&0.37&0.47&0.57&0.71&0.26&0.48&0.74&0.95&0.96&0.31&0.66&0.88&0.96&1.05\\
\hline
SkelNet(Ours)&0.23&0.39&0.50&0.58&0.71&
0.26&0.47&\textbf{0.70}&\textbf{0.91}&\textbf{0.89}&
\textbf{0.29}&\textbf{0.62}&0.88&\textbf{0.89}&1.00\\
Skel-TNet(Ours)&\textbf{0.21}&\textbf{0.34}&\textbf{0.43}&\textbf{0.55}&\textbf{0.70}&\textbf{0.25}&\textbf{0.46}&0.71&\textbf{0.91}&0.90&0.30&0.64&\textbf{0.86} &0.90&\textbf{0.99}\\
\hline
\end{tabular}
}
\caption{ Detailed results reported in Euler angle error on representative activities of the Human3.6M dataset, for short-term (80, 160, 240, 320 and 400ms) prediction. Bold means the best performance.}
\label{3_activities}
\end{table*}
\textbf{Shot-term motion prediction: } In this task, we firstly train Skel-TNet by minimzing the prediction error over future 400 ms. We report results measured by MoF for every single activities in two datasets in Table \ref{short_term_prediction}. We can notice that Skel-TNet surpasses or at least catches up with CSS, specifically, results are more accurate: lowering averaging errors of the second best results on two datasets by 5\% and 10\%, respectively. Although SkelNet is designed for long-term prediction task, we also train it for short-term prediction task, and it performs comparably with the state-of-the-art method on many activity predictions, and it achieves slightly lower averaging MoF than CSS.\\
\indent In Table \ref{3_activities}, we compare with multiple networks for human motion prediction on several time-steps (i.e. 80, 160, 240, 320 ms). We can notice that the best result constantly fallen into one of our methods. Albeit slight increasing margins in Table \ref{3_activities}, we still believe that these margins can be enlarged by extending Skel-TNet's components to more complicated ones, which are with higher learning capability regarding each spatial and temporal dependencies.
\subsection{Ablative study}
\begin{table}[ht]
\centering
\resizebox{0.5\textwidth}{!}{
\begin{tabular}{ccc}
& Human3.6M& CMU mocap dataset\\
\hline
full &\textbf{1.17}&0.89\\
\hline
w long prior&1.19&\textbf{0.87}\\
\hline
w tanh&1.19&0.92\\
w/o residual connection&2.48&1.77\\
w/o brances&1.23&0.94\\
w/o branches w/o LReLUs&1.47&1.08\\
w/o branches w/o LReLUs w/o dropout&1.53&1.16\\
\hline
SkelNet\textunderscore UD&1.19&0.93\\
SKelNet\textunderscore LR&1.21&0.94\\
\hline
\end{tabular}
}
\caption{Ablation studies on different components in the network design of our SkelNet, results reported is the mean value of MoF for all activities.}
\label{t_skelNet_archi}
\end{table}
\begin{table}[ht]
\centering
\resizebox{0.5\textwidth}{!}{
\begin{tabular}{c| ccc| ccc| c}
\hline
&\multicolumn{3}{c|}{Human3.6M}&\multicolumn{3}{c|}{CMU mocap dataset}& \\
Noise Variance&0.1&0.3&0.5&0.1&0.3&0.5&Total parameters\\
RRNN&1.65&1.74&1.87&&&& $\approx 0.5m$ \\
CSS&1.26&\textbf{1.50}&1.82&1.03&1.21&1.50& $\approx 9m$ \\
SkelNet&\textbf{1.24}&\textbf{1.50}&\textbf{1.80}&\textbf{0.93}&\textbf{1.15}&\textbf{1.41}& $\approx 0.3m$\\
\hline
\end{tabular}
}
\caption{The averaging prediction error reported in MoF on two datasets, the input is under Gaussian noise with different variance [0.1, 0.3, 0.5]. The rightmost column records total number of parameters used in each method.}
\label{noise_resiliency}
\end{table}
\textbf{SkelNet:} We firstly study the effectiveness of the baseline, a standard feed-forward network for predicting human motion over long-term, by removing LReLUs, dropout and branches (w/o branches w/o LReLUs w/o dropout), this cause the error to increase; its performance can be improved by adding the dropout (w/o branches w/o LReLUs) or adding both the dropout and LReLUs (w/o branches). Furthermore, we illustrate the difference between full SkelNet (full) with or without branches (w/o branches), the difference indicates that branches in SkelNet do improve prediction performance, we think this is because intra-pose interventions can be largely tackled. This idea is further supported by Figure \ref{Skel_Plain}, except Eular angle error on left arm, errors on other components and complete human pose are reduced. Moreover, our SkelNet's effectiveness will not be restricted when long prior is available, as shown by the first two rows. \\
\indent Furthermore, we compare with two networks with 3 branches of layers, using two different groupings: left (leg+arm), torso, right(leg+arm), denoted as Skel\textunderscore LR; (left+right) arms, torso, (left+right) legs, denoted as Skel\textunderscore UD; Performance of Skel\textunderscore LR and Skel\textunderscore UD are both worse than original SkelNet with five branches of layers that better capture component-specific information.\\
\indent Also, we investigate the robustness of SkelNet. We train SkelNet on long-term prediction, input data in both training and testing is under Gaussian noise with different variances ([0.1, 0.3, 0.5]). Table \ref{noise_resiliency} shows that SkelNet has advantage in terms of prediction accuracy to two competitors in the CMU mocap dataset, no matter what variance of Gaussian noise has been used. However, it is only able to perform comparably with CSS on the Human3.6M dataset. Margins in both datasets between SkelNet and CSS are not increased as variance goes up. However, safe conclusion can be drawn that SkelNet is slightly more robust than its competitors. \\
\begin{table}[t]
\centering
\resizebox{0.45\textwidth}{!}{
\begin{tabular}{c|cccc|cccc}
& \multicolumn{4}{c}{Human3.6M} & \multicolumn{4}{c}{CMU mocap dataset} \\
\hline
&Walk&Eat&Discuss&Avg&Run&BSignal&Soccer&Avg \\
SkelNet&\textbf{0.49}&0.46&\textbf{0.70}&\textbf{0.76}&\textbf{0.49}&0.38&\textbf{0.48}&0.60\\
C-RNN&0.55&\textbf{0.41}&\textbf{0.70}&\textbf{0.76}&0.59&\textbf{0.27}&0.52&\textbf{0.59}\\
C-RNN(w/o)&0.56&0.45&0.85&0.79&0.59&0.33&0.64&0.63\\
\hline
\end{tabular}
}
\caption{Ablation studies on Skel-TNet's components. All MoF reported are for specific activities and averaging value computed on all activities of two datasets. C-RNN and C-RNN(w/o) denote C-RNN trained by  $\mathcal{L}_{conv}$ and \textit{Sampling-based loss}.}
\label{t_skelNet_component}
\end{table}
\begin{table}[t]
\centering
\resizebox{0.45\textwidth}{!}{
\begin{tabular}{ccc}
& Human3.6M& CMU mocap dataset \\
\hline
Full & \textbf{0.73}&\textbf{0.55}\\
\hline
w/o T&0.78&0.60\\
w/o S&0.79&0.59\\
w/o M&0.80&0.62\\
\hline
Trained jointly(1)&0.82&0.69\\
Trained jointly(2)&0.83&0.67\\
\hline
\end{tabular}
}
\caption{Ablation studies on Skel-TNet's architecture. Results reported is the mean value of MoF for all activities. Trained jointly (1) and (2) represent arranging Skel-TNet alternatives in two different orders.}
\label{t_skelTNet_archi}
\end{table}

\textbf{Skel-TNet component:} We exam the performance of Skel-TNet's components, training them independently for short-term prediction and results are reported in Table \ref{t_skelNet_component}. Generally, SkelNet and C-RNN are able to produce predictions with roughly same MoF in averaging results computed over all activities. However, SkelNet has advantages to C-RNN on activities like ``WashWindow'' and ``Walking'', conversely, C-RNN achieve the more accurate prediction on ``BasketballSignal'' and ``Eating''. Moreover, our $\mathcal{L}_{conv}$ do provide C-RNN with improvement in its generalization capability, as shown in prediction results on averaging MoF over all activities and MoF on some certain activities (``Soccer'' and ``Discussion'').\\

\textbf{Skel-TNet architecture:} The Skel-TNet is a multi-staged processing network, there are several alternatives to its configuration. We firstly test Skel-TNet's generalization ability when Merging Network is merely with one component, SkelNet (w/o T) or C-RNN (w/o S), by removing the other. Furthermore, we remain both of SkelNet and C-RNN but replace Merging Network with averaging sum for two input sequences (w/o M). At the end, we train three components in stacked way, orders are SkelNet, C-RNN and Merging Network, or C-RNN, SkelNet and Merging Network. We get rid of one of two trainable weights in Merging Network applied on input sequences, when there is only one sequence being input to Merging Network. All results are in Table \ref{t_skelTNet_archi}.\\
\section{Conclusion}
We propose SkelNet for human motion prediction at first. Unlike the previous work, we pay attention to different dynamic patterns from local components of the human pose, this is achieved by dividing the human pose into five parts that then be fed into five component-specific layers of SkelNet for obtaining representations of local structures. Our SkelNet is designed as effective yet simple (1/30 parameter number of the state-of-the-art method) and can be strengthened for its capability on learning temporal dynamics in the Skel-TNet, which is our the second proposed network for prediction task. Both SkelNet and Skel-TNet obtain superior or at least comparable performances to recently proposed methods, as shown by experimental results.
\bibliographystyle{aaai}
\bibliography{References_}

\end{document}